\documentclass[final]{cvpr}

\usepackage{times}
\usepackage{epsfig}
\usepackage{graphicx}
\usepackage{amsmath}
\usepackage{amssymb}

\usepackage{subfigure}
\usepackage{lipsum}
\usepackage{multirow}

\usepackage{booktabs,tabularx}
\usepackage{siunitx}
\newcolumntype{C}{>{\centering\arraybackslash}X}

\usepackage[pagebackref=true,breaklinks=true,colorlinks,bookmarks=false]{hyperref}

\def\cvprPaperID{****} % *** Enter the CVPR Paper ID here
\def\confYear{CVPR 2021}

\begin{document}

%%%%%%%%% TITLE
\title{Boosting Video Representation Learning with Multi-Faceted Integration}

\author{Zhaofan Qiu$^{\dagger}$, Ting Yao$^{\dagger}$, Chong-Wah Ngo$^{\ddagger}$, Xiao-Ping Zhang$^{\S}$, Dong Wu$^{\dagger}$ and Tao Mei$^{\ddagger}$\\
\parbox{20em}{\small\centering $^{\dagger}$ JD AI Research, Beijing, China}\\
\parbox{40em}{\small\centering $^{\ddagger}$ Singapore Management University, Singapore~~~~~~~~~~~~~~~~~~~$^{\S}$ Ryerson University, Toronto, Canada}\\
{\tt\small \{zhaofanqiu, tingyao.ustc\}@gmail.com, cwngo@smu.edu.sg} \\
{\tt\small xzhang@ee.ryerson.ca, wudong99@gmail.com, tmei@jd.com}
}

\maketitle
\thispagestyle{empty}

%%%%%%%%% ABSTRACT
\begin{abstract}
Video content is multifaceted, consisting of objects, scenes, interactions or actions. The existing datasets mostly label only one of the facets for model training, resulting in the video representation that biases to only one facet depending on the training dataset. There is no study yet on how to learn a video representation from multifaceted labels, and whether multifaceted information is helpful for video representation learning. In this paper, we propose a new learning framework, MUlti-Faceted Integration (MUFI), to aggregate facets from different datasets for learning a representation that could reflect the full spectrum of video content. Technically, MUFI formulates the problem as visual-semantic embedding learning, which explicitly maps video representation into a rich semantic embedding space, and jointly optimizes video representation from two perspectives. One is to capitalize on the intra-facet supervision between each video and its own label descriptions, and the second predicts the ``semantic representation'' of each video from the facets of other datasets as the inter-facet supervision. Extensive experiments demonstrate that learning 3D CNN via our MUFI framework on a union of four large-scale video datasets plus two image datasets leads to superior capability of video representation. The pre-learnt 3D CNN with MUFI also shows clear improvements over other approaches on several downstream video applications. More remarkably, MUFI achieves 98.1\%/80.9\% on UCF101/HMDB51 for action recognition and 101.5\% in terms of CIDEr-D score on MSVD for video captioning.
\end{abstract}
 
\section{Introduction}
Deep Neural Networks have been proven to be highly effective for learning vision models on large-scale datasets. To date in the literature, there are various image datasets (e.g., ImageNet \cite{russakovsky2015imagenet}, COCO \cite{lin2014microsoft}, Visual Genome \cite{krishna2017visual}) that include large amounts of expert labeled images for training deep models. The well-trained models, on one hand, manifest impressive classification performances, and on the other, produce discriminative and generic representation for image understanding tasks. Compared to static 2D images, video has one more dimension (time) and is an information-intensive media with large variations and complexities. As a result, learning a powerful spatio-temporal video representation is yet a challenging problem.
 
\begin{figure}[!tb]
   \centering {\includegraphics[width=0.40\textwidth]{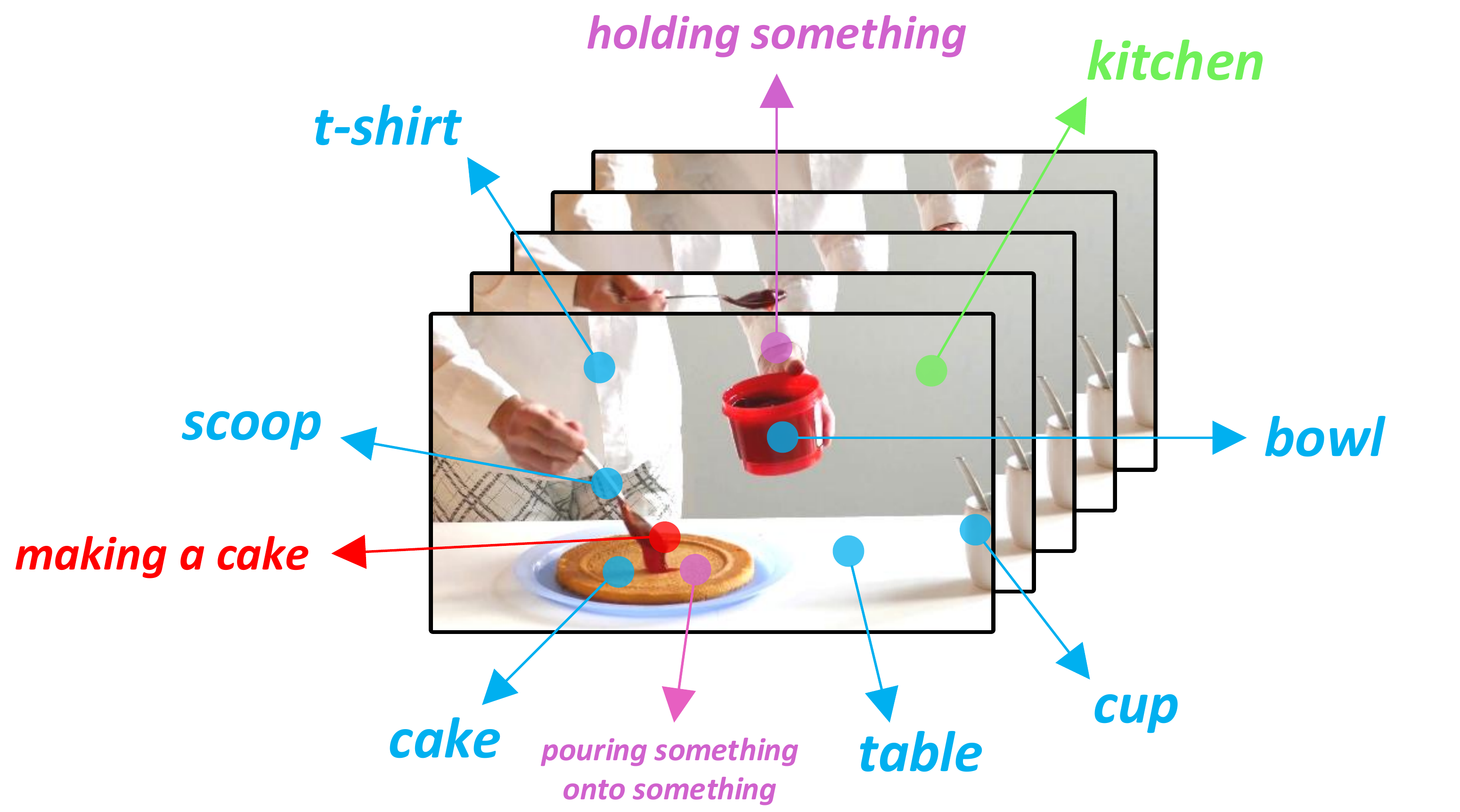}}
   \caption{\small Example illustrating the four facets of video content: object (blue), scene (green), interaction (magenta), and action (red).}
   \label{fig:intro}
   \vspace{-0.25in}
\end{figure}
 
Capitalizing on the high capability of deep models, one natural way to improve video representation is to acquire more video data. For example, Tran \emph{et al.} \cite{tran2015learning} devised a widely-adopted 3D CNN, namely C3D, optimized on a large-scale Sports1M dataset and a constructed I380K dataset. Carreira \emph{et al.} \cite{carreira2017quo} built a popular pre-training dataset, i.e., Kinetics, consisting of around 300K well-annotated trimmed video clips. To expand the study in the regime of web videos which is multiple orders of magnitude larger, Ghadiyaram \emph{et al.} \cite{ghadiyaram2019large} collected 65M web videos for pre-training 3D CNN in a weakly-supervised manner. Despite the tremendous progresses, performing learning on a specific dataset usually focuses on a particular channel of videos (e.g., action) and seldom explores other facets of videos simultaneously. Taking a video of \emph{``making a cake''} from Kinetics dataset as an example (Figure \ref{fig:intro}), there are a wide variety of facets, ranging from object, scene, interaction, to action. Nevertheless, Kinetics, as human action dataset, mainly emphasizes the facet of action, making the learnt representation mostly aware of the action information. A valid question is how to leverage multifaceted video information for representation learning.

This paper explores the integration of six facets, i.e., action (Kinetics \cite{carreira2017quo}), event (Moments-In-Time \cite{monfort2019moments}), interaction (Something-Something \cite{Goyal2017TheS}), sport (Sports1M \cite{ghadiyaram2019large}), object (ImageNet \cite{russakovsky2015imagenet}), and scene (Place365 \cite{zhou2017places}), and each facet corresponds to one dataset. We aim for a model that engages all the six facets (datasets) to learn video representation, ideally making the representation more discriminative and generic. The inherent difficulty of learning such a representation is: \emph{how to execute effective representation learning on various datasets with different labels in a unified framework?} We propose to mitigate this issue through visual-semantic embedding learning. The basic idea is to learn a semantic space that bridges the labels from different datasets. The model that generates the space is pre-trained on a large-scale unannotated text data. The learning objective is to model the semantic relationships between labels and embed the disjoint labels into semantically-meaningful feature vectors. We project video representation into the semantic space and optimize visual-semantic embedding to enhance representation learning.
 
To materialize our idea, we present a new MUlti-Faceted Integration (MUFI) framework for video representation learning. Specifically, we employ off-the-shelf language models such as BERT \cite{devlin2018bert} to extract textual features and build the semantic space, which is also taken as the embedding space. Each video is fed into a 3D CNN to obtain the video representation and then mapped into the embedding space. The learning of visual-semantic embedding is supervised by intra-facet embedding of a video and its class label within a dataset, and inter-facet label prediction and embedding across multiple datasets. Our MUFI framework capitalizes on the two types of supervision to jointly learn visual-semantic embedding and adjust 3D CNN through a multi-attention projection structure, and performs the whole training in an end-to-end manner. Note that the image datasets only offer the inter-facet supervision to the videos here and are not exploited as the network inputs.
 
The main contribution of this work is the exploration of multifaceted video content from various datasets to improve video representation learning. The novel idea leads to the elegant views of how to relate the facets of videos from different datasets, and how to consolidate various facets in a unified framework for learning, which are problems not yet fully understood. We demonstrate the effectiveness of our MUFI framework on a union of four large-scale video datasets plus two image datasets in the experiments.

\section{Related Work}
The early works of using Convolutional Neural Networks for video representation learning are mostly extended from 2D CNN for image classification \cite{diba2017deep,feichtenhofer2016convolutional,karpathy2014large,qiu2017deep,simonyan2014two,wang2016temporal,wang2018temporal,zhu2016key}. These approaches often treat a video as a sequence of frames or optical flow images, while overlooking the pixel-level temporal evolution across consecutive frames. To alleviate this issue, 3D CNN \cite{ji20133d,tran2015learning} is devised to directly learn spatio-temporal representation from a short video clip via 3D convolution, which shows good transferability to several downstream tasks \cite{li2018recurrent,li2019long,long2019gaussian,long2020learning,long2019coarse,qiu2017learning2}. Despite having encouraging performances, the training of 3D CNN is computationally expensive and the model size suffers from a massive growth. Later in \cite{qiu2017learning,tran2018closer,xie2018rethinking}, the 3D convolution is approximately decomposed into one 2D spatial convolution plus one 1D temporal convolution. Recently, more advanced techniques are presented for 3D CNN, including inflating 2D kernels \cite{carreira2017quo}, non-local pooling \cite{wang2018non}, local-and-global diffusion \cite{qiu2019learning}, and filter banks \cite{Martnez2019ActionRW}.

The aforementioned works predominately focus on the designs of network architectures, and another direction of video representation learning is to involve more data. Ghadiyaram \emph{et al.} \cite{ghadiyaram2019large} collect a large dataset with 65M web videos, which is multiple orders of magnitude larger than the existing video datasets. The 3D CNN in \cite{ghadiyaram2019large} is trained on such large dataset in a weakly-supervised manner and shows obvious improvements over other pre-training strategies. Later, OmniSource \cite{duan2020omni} further improves the work by utilizing weakly-labeled web images and additional web videos. Instead of treating search queries as weakly-supervised labels, the title of video in the website can also be regarded as weak supervision. In \cite{li2020learning}, the correlation between a video and its associated title is explored to learn a rich semantic video representation. Inspired by the recent advances of self-supervised learning, there are also some approaches \cite{qian2020spatiotemporal,wang2020self,yang2020video,yao2020seco} which propose to learn video representation from unannotated video data.

In summary, our work belongs to supervised video representation learning. Different from the existing methods that focus learning on a particular facet of videos, our approach contributes by studying how to present the facets of videos from different datasets and proposing a novel video representation learning framework to consolidate various facets.

\begin{figure*}[!tb]
   \centering
   \subfigure[\scriptsize Classification]{
     \label{fig:schematic:a}
     \includegraphics[width=0.11\textwidth]{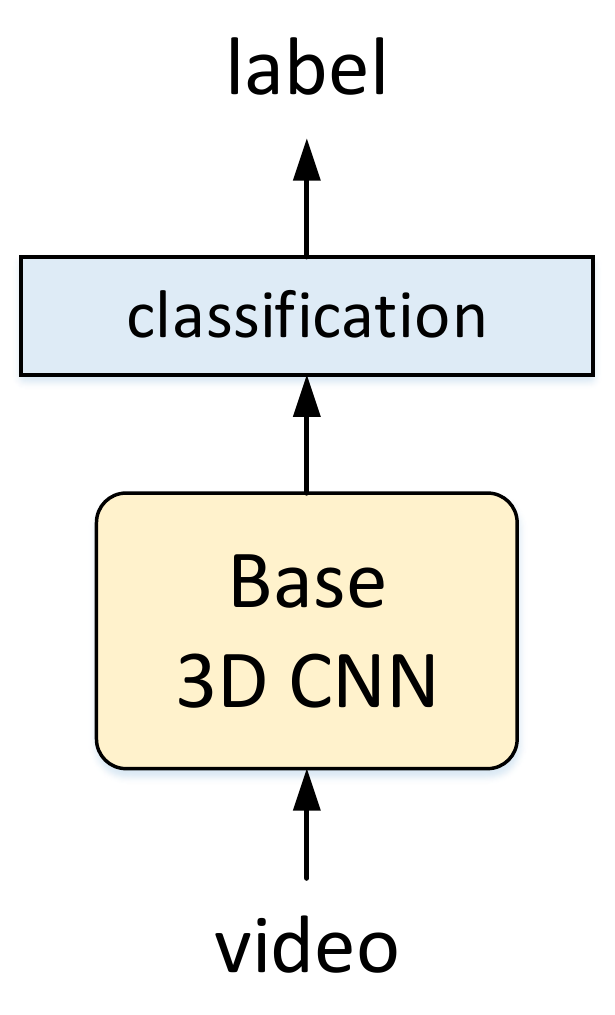}}
\hspace{0.1in}
   \subfigure[\scriptsize Visual-Semantic Embedding]{
     \label{fig:schematic:b}
     \includegraphics[width=0.185\textwidth]{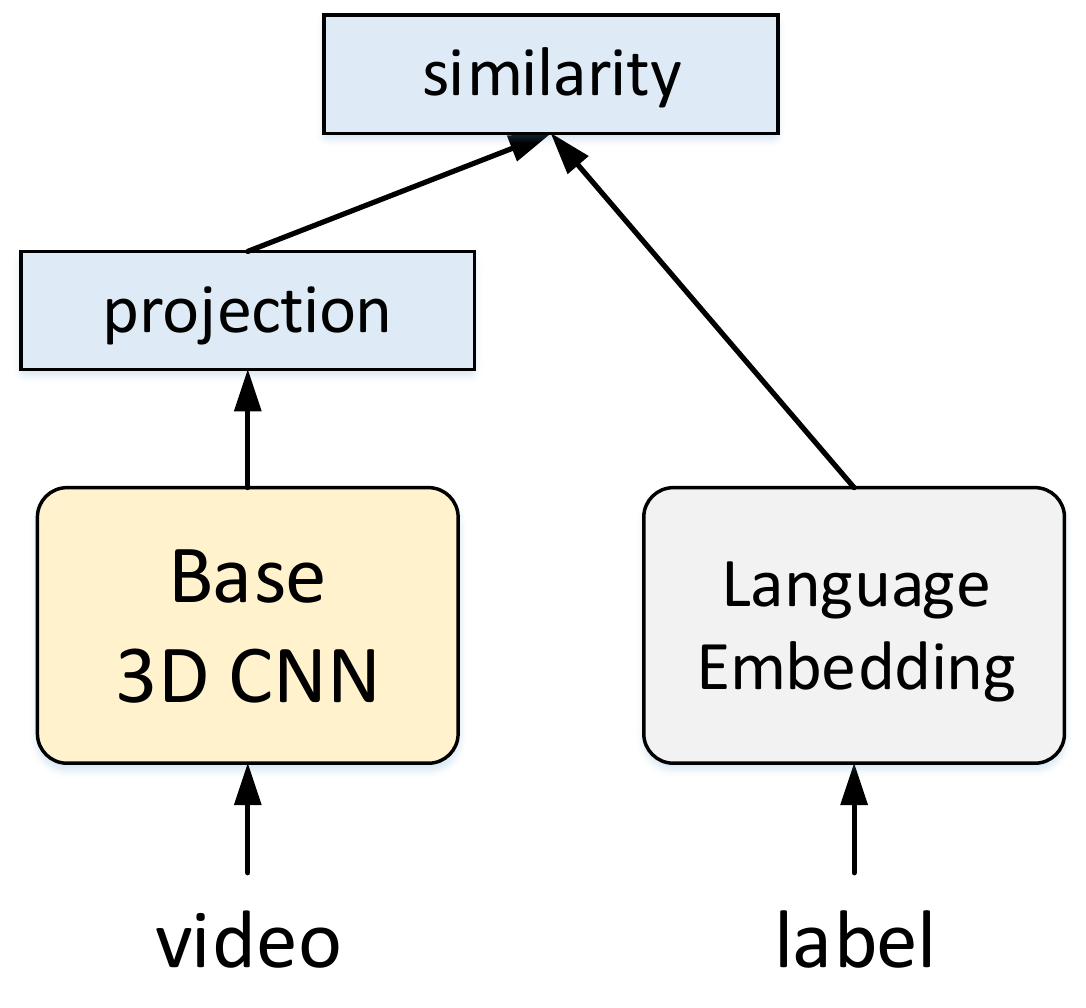}}
\hspace{0.1in}
   \subfigure[\scriptsize Intra-Facet Supervision Only]{
     \label{fig:schematic:c}
     \includegraphics[width=0.20\textwidth]{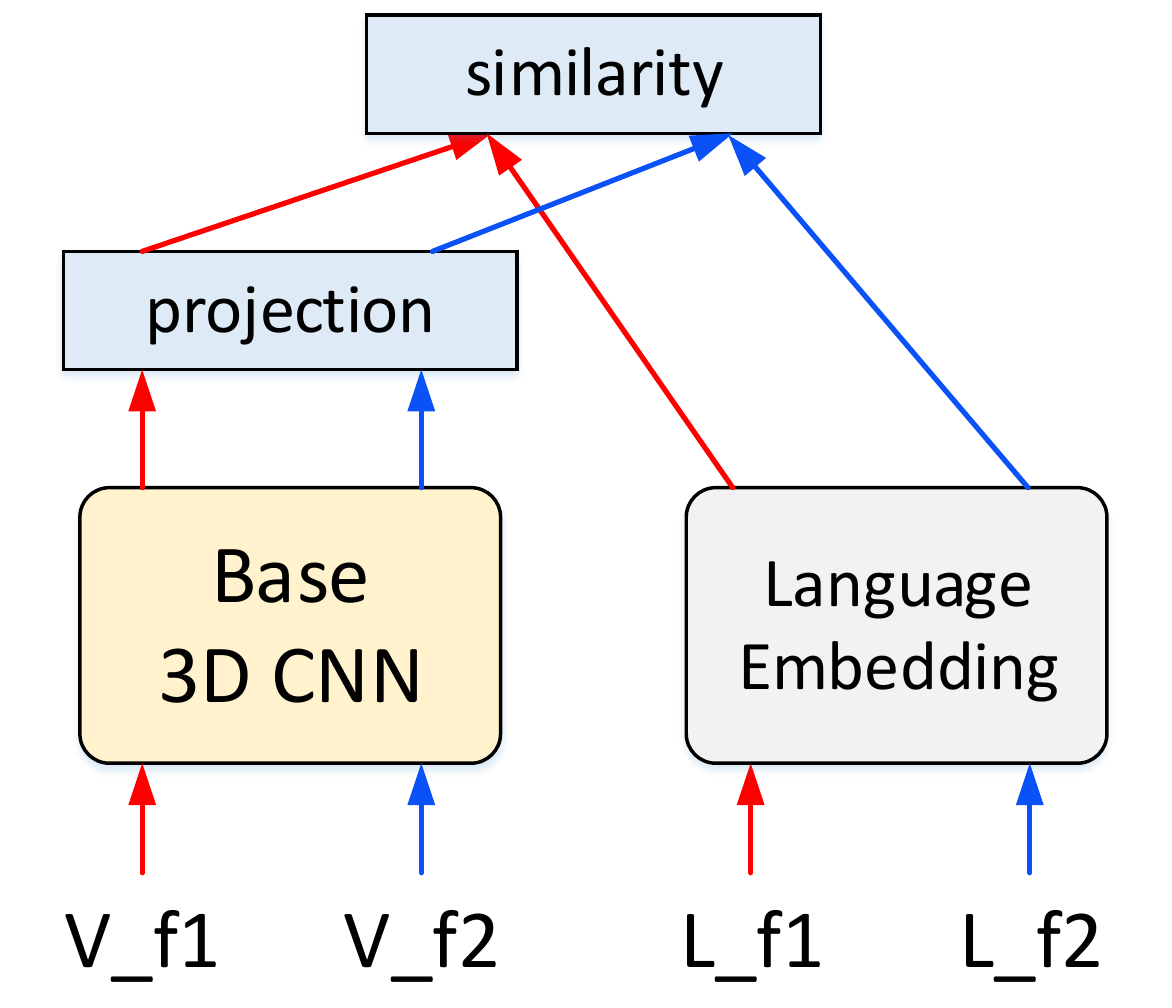}}
\hspace{0.1in}
   \subfigure[\scriptsize Intra-Facet Supervision + Inter-Facet Supervision]{
     \label{fig:schematic:d}
     \includegraphics[width=0.40\textwidth]{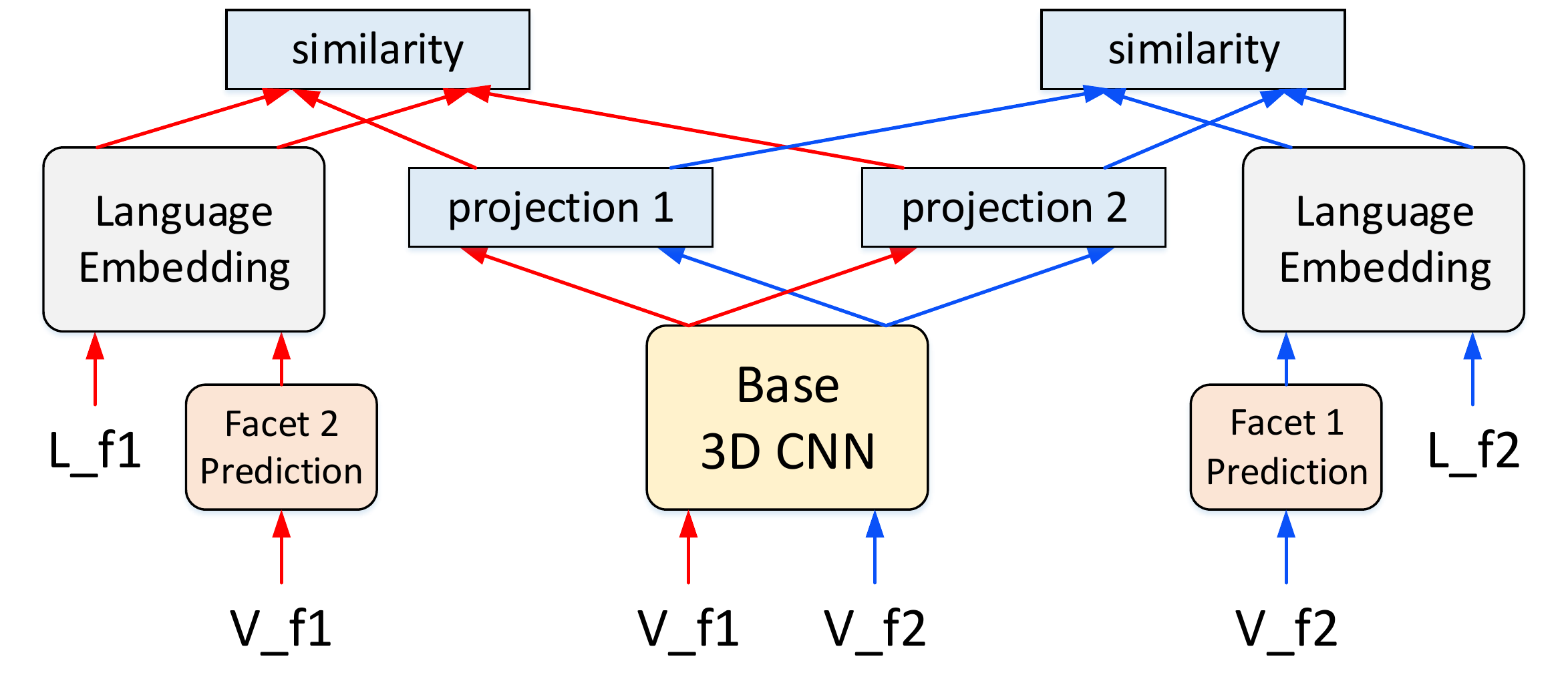}}
   \caption{\small The schematic illustration of video representation learning by (a) classification loss; (b) visual-semantic embedding; (c) multi-faceted visual-semantic embedding with intra-facet supervision; (d) multi-faceted visual-semantic embedding with intra-facet supervision and inter-facet supervision. For the multi-faceted frameworks, the arrows of data flow from an identical facet are in the same color.}
   \label{fig:schematic}
   \vspace{-0.20in}
\end{figure*}

\section{Multi-Faceted Integration}

In this paper, we devise MUlti-Faceted Integration (MUFI) framework to integrate the rich facets from different datasets to boost video representation learning. Specifically, MUFI formulates the problem as visual-semantic embedding learning, which explicitly maps video representation into a rich semantic embedding space. We begin this section by introducing the visual-semantic embedding for video representation learning, followed by the two learning perspectives, i.e., intra-facet supervision and inter-facet supervision. Then, we propose a multi-attention projection structure to embed the visual feature. Finally, we elaborate the end-to-end optimization of our MUFI.

\subsection{From Classification to Embedding}

We firstly study the most common single-facet representation learning problem. Suppose we have a labeled video collection $\mathcal{V}=\{(v, l)\}$, where $l \in \mathcal{L}$ is the assigned label to describe the video content. The goal is to pre-train a visual encoder $\mathcal{F}(\cdot)$ on video data supervised by the semantic labels. The pre-trained video encoder can be further utilized to support several video downstream tasks. One conventional way is to optimize the encoder $\mathcal{F}(\cdot)$ with classification loss, as shown in Figure \ref{fig:schematic:a}. In our case, we choose 3D CNN as visual encoder and then append a linear classifier $\mathcal{C(\cdot)}$ on top of the network to predict the label. Hence, the visual encoder and classifier are jointly optimized by Softmax Cross-Entropy (SCE) loss as
\begin{equation}\label{eq:classification}
\begin{aligned}
L_{classification}(v, l)=\text{SCE}\left(\mathcal{C}(\mathcal{F}(v)), l\right).
\end{aligned}
\end{equation}

Instead, the multi-faceted learning involves $N$ video collections $\mathcal{V}^n|_{n=1,...,N}$ and the corresponding label set $\mathcal{L}^n$. Eq.(\ref{eq:classification}) can be straightforwardly extended by merging the label sets or establishing multiple classifiers. Nevertheless, these simple extensions treat each facet independently, but ignore the semantic relationships between labels.

Inspired by deep visual-semantic embedding \cite{frome2013devise}, learning is formulated as a problem of aligning video representation and label representation in a rich semantic embedding space, as shown in Figure \ref{fig:schematic:b}. Formally, the label representation is achieved by a frozen language embedding model $\mathcal{S(\cdot)}$ (e.g., BERT) pre-trained on unannotated text data. The objective of visual-semantic embedding is to minimize the distance between the projected video representation and the label representation:
\begin{equation}\label{eq:embedding}
\begin{aligned}
L_{embedding}(v, l)=\text{distance}\left(\mathcal{P}(\mathcal{F}(v)), \mathcal{S}(l)\right),
\end{aligned}
\end{equation}
where $\mathcal{P}(\cdot)$ is a projection function to embed the video representation into the semantic space. As such, the labels from different datasets share the same continuous feature space and are comparable to each other.

\subsection{Intra-Facet Supervision}

To tackle the challenge of multi-faceted embedding, we first introduce an intra-facet supervision that minimizes the distance between video representation and label representation in the same facet, as illustrated in Figure \ref{fig:schematic:c}. Given a video-label pair $(v^n, l^n)$ from the $n^{th}$ facet, the intra-facet supervision can be formulated as minimizing the L2 distance between embedded representations:
\begin{equation}\label{eq:intra-l2}
\begin{aligned}
L_{intra-l2}(v^n, l^n)=\left \|  \mathcal{P}(\mathcal{F}(v^n)) - S(l^n)\right \|_{2}^{2}.
\end{aligned}
\end{equation}

However, minimizing the L2 loss only tends to make the video representation close to the annotated label while ignoring other labels in the same facet. Such kind of training may affect the discriminative capability of the learnt representation in this facet, since other labels might be even closer than the annotated label. Similar phenomenon is also discussed in \cite{frome2013devise}. Here, we propose to address this issue by the recent success of contrastive learning \cite{gutmann2010noise,hadsell2006dimensionality,oord2018representation}. The basic principle is to make positive/negative query-key pairs similar/dissimilar. In our case, the positive pair is constructed by taking the embedded visual representation $\mathcal{P}(\mathcal{F}(v^n))$ as query and the representation of the annotated label $S(l^n)$ as positive key. The representations from other labels in the same facet are regarded as the negative keys. Formally, by measuring the query-key similarity via dot product, the intra-facet loss can be formulated based on a softmax formulation:
\begin{equation}\label{eq:intra-nce}
\begin{aligned}
L_{intra-nce}(v^n, l^n)=-\text{log}\frac{\text{exp}(\mathcal{P}(\mathcal{F}(v^n))\cdot S(l^n))}{\sum\limits_{\hat{l}^n \in \mathcal{L}^n} \text{exp}(\mathcal{P}(\mathcal{F}(v^n))\cdot S(\hat{l}^n))},
\end{aligned}
\end{equation}
which is similar as the prevailing form of contrastive loss in InfoNCE \cite{oord2018representation} but without the temperature parameter. The motivation behind Eq.(\ref{eq:intra-nce}) is to minimize the distance between a query (video representation) and its positive key (the assigned label) while remaining to be distinct to negative keys (the other labels) in the meantime.

\subsection{Inter-Facet Supervision}

In the intra-facet supervision, each video is only supervised by a single label from one facet (its source facet), while leaving the other information in the video unexploited.
Although we can pool the multiple datasets together by leveraging the single-faceted labels of a dataset at a time, learning is still based on single facet and multifaceted analysis is not exploited, resulting in data inefficiency. To alleviate the issue, we additionally involve the inter-facet supervision on video data to explore the potential of the other facets via transferring the pre-learnt knowledge, as illustrated in Figure \ref{fig:schematic:d}. Specifically, when transferring the knowledge of the $m^{th}$ facet to video $v^{n}\in \mathcal{V}^n$, we utilize a classification model pre-trained on the $m^{th}$ facet to predict the category probability of the input video as $p^m(\hat{l}^m|v^n)$. Our goal is to make the embedded video representation close to the category with high probability in the $m^{th}$ facet. Thus, we calculate the sum of label representations weighted by their label probabilities as the pseudo ``semantic representation'' of video $v^n$ as
\begin{equation}\label{eq:transfer}
\begin{aligned}
S^m(v^n) = \sum\limits_{\hat{l}^m \in \mathcal{L}^m}p^m(\hat{l}^m|v^n) S(\hat{l}^m).
\end{aligned}
\end{equation}
The classification model can be considered as an expert ``labeler'' in $m^{th}$ facet, and the prediction from this ``labeler'' is mapped to the semantic space. Then, the objective of inter-facet supervision is to minimize the distance between embedded video representation and pseudo representation:
\begin{equation}\label{eq:inter}
\begin{aligned}
L_{inter}(v^n) = \sum\limits_{m\neq n} \left \|  \mathcal{P}^m(\mathcal{F}(v^n)) -S^m(v^n)\right \|_{2}^{2}.
\end{aligned}
\end{equation}
Unlike intra-facet supervision with strict labels, we do not require the embedded visual feature to become far apart from the other labels for inter-facet supervision and thus simply choose the L2 loss. By minimizing the loss in Eq.(\ref{eq:inter}), the videos from any dataset can be supervised by all the facets. Moreover, our proposal also provides an elegant way to learn from image datasets by transferring the pre-trained image classifier to video data.
\begin{figure}[!tb]
   \centering {\includegraphics[width=0.40\textwidth]{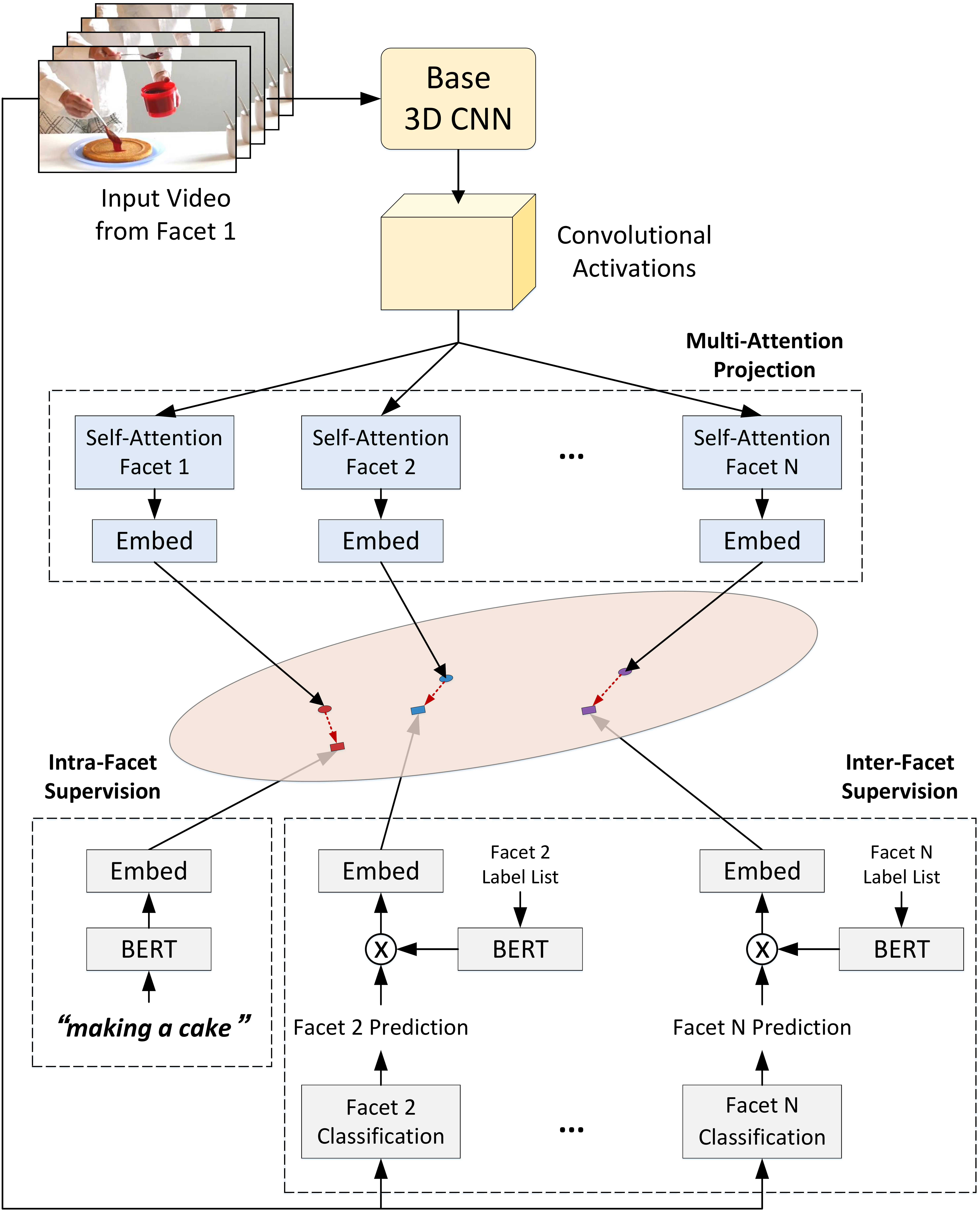}}
   \caption{\small The overview of our MUFI framework for video representation learning. In the figure, we take the video of \emph{``making a cake''} from facet $1$ as an example. The video feature extracted from a base 3D CNN is mapped into a semantic embedding space by the multi-attention projection. The base 3D CNN and projection function is optimized in an end-to-end manner with both the intra-facet supervision and inter-facet supervision.}
   \label{fig:framework}
   \vspace{-0.20in}
\end{figure}

\subsection{Multi-Attention Projection}
In many cases, each facet is only related to some local regions that contain special objects or motion. Therefore, directly using one global feature (e.g., feature from global pooling layer) may lead to suboptimal results due to the noises introduced from regions that are irrelevant to the facet. In order to enable the framework to pinpoint the spatio-temporal regions that are most indicative for each facet, we devise a self-attention component plus a linear embedding layer as the projection function $\mathcal{P}^n$ for each facet. The self-attention assigns a positive weight score to each local descriptor extracted from the last convolutional layer of the 3D CNN. The score can be interpreted as the probability that the spatio-temporal region should be attended for this facet. Then the linear embedding layer maps the attended feature to the semantic space.

Technically, given the feature map $\mathcal{F}(v) \in \mathbb{R}^{T\times H\times W}$ extracted from the 3D CNN, the weight-sum of local descriptors by self-attention can be calculated by
\begin{equation}\label{eq:attend}
\begin{aligned}
\sum\limits_{f_{(i)} \in \mathcal{F}(v)}\varphi(w^n \cdot f_{(i)} + b^n) f_{(i)},
\end{aligned}
\end{equation}
where $\varphi(\cdot)$ is the softmax normalization across different spatio-temporal positions to produce the positive attention probabilities in the range of $(0, 1)$. $\{w^n, b^n\}$ are learnable parameters to distinguish the important regions for the $n^{th}$ facet. Please note that more complex projection functions, e.g., Multiple Layer Perception (MLP), could improve the embedding performance but meanwhile might influence the learnt representation. In view that our goal is to train the base 3D CNN for a better video representation, we choose the relatively simple multi-attention projection.

\subsection{Overall Framework}
Figure \ref{fig:framework} shows the overall framework of our MUlti-Faceted Integration (MUFI), which utilizes the proposed intra-facet supervision, inter-facet supervision and multi-attention projection. Considering that the ultimate goal of MUFI is to train the base 3D CNN for video representation, we fix the pre-trained language model, embedding layer for language feature and the classification models, and only optimize the base 3D CNN and the multi-attention projection. Specifically, the language model is BERT pre-trained on large-scale unannotated text data. The parameters of language embedding layer is achieved by Principal Component Analysis (PCA) across all the label representations. For the base 3D CNN and the multi-attention projection, we update the parameters according to the following overall loss as
\begin{equation}\label{eq:overall}
\begin{aligned}
L(v^n, l^n) = L_{intra-nce}(v^n, l^n) + L_{inter}(v^n).
\end{aligned}
\end{equation}
Here, we empirically treat each facet in MUFI equally.

\section{Experiments}

\subsection{Dataset and Evaluation}
We train our MUFI framework on a union of six large-scale datasets: Kinetics \cite{carreira2017quo}, Moments-In-Time \cite{monfort2019moments}, Something-Something \cite{Goyal2017TheS}, Sports1M \cite{ghadiyaram2019large}, ImageNet \cite{russakovsky2015imagenet} and Place365 \cite{zhou2017places}. The first four are video recognition datasets focusing on action, event, interaction and sport, respectively. The last two are popular image datasets for object recognition and scene recognition.

\textbf{Kinetics} is one of the large-scale video benchmarks for action recognition on trimmed video clips. We utilize the latest released version of the dataset, i.e., Kinetics-700, which contains around 600K video clips from 700 action categories. \textbf{Moments-In-Time} is a large-scale video event recognition dataset with around 800K video moments (3-second short videos) from 339 events. The \textbf{Something-Something} dataset is firstly constructed in \cite{Goyal2017TheS} to learn fine-grained human-object interaction, and then extended to Something-Something V2 recently. \textbf{Sports1M } dataset consists of about 1.13M videos annotated with 487 sports labels. Please note that some video URLs were dead when we downloaded the videos. Hence, we only employ around 1M available videos. For each raw video, we randomly sample five 5-second short clips for efficient training. \textbf{ImageNet} and \textbf{Place365} are image benchmarks with over 1M images from 1,000 object categories and 365 scene categories, respectively. These two datasets are only exploited to provide inter-facet supervision and never used as inputs to 3D CNN.

\textbf{Per-facet evaluation.} To thoroughly evaluate the learnt representation with respect to every facet, we construct a multifaceted evaluation metric on the classification of action (\textbf{act.}), event (\textbf{eve.}), interaction (\textbf{int.}), sport (\textbf{spo.}), object (\textbf{obj.}) and scene (\textbf{sce.}), respectively. For the first four facets, we randomly sample 50 videos of each class from the training/validation set of Kinetics-700, Moments-In-Time, Something-Something V2 and Sports1M datasets as the standalone training/validation set. The videos in these four subsets will be only exploited to evaluate the learnt representation and are not used to train MUFI. For object recognition, we utilize the \textbf{ImageNet object detection from video} (VID) \cite{russakovsky2015imagenet} dataset with 3,862 video clips of 30 object categories. Please note that VID dataset provides the bounding box of each object and we only use the most frequent object as the video label. For scene recognition, we employ \textbf{YUP++} \cite{feichtenhofer2017temporal} that contains 1,200 video from 20 scene categories.

\textbf{Downstream task evaluation.} We also verify the merit of MUFI in three downstream tasks: action recognition (UCF101 \cite{UCF101}, HMDB51 \cite{HMDB51} and Kinetics-400), interaction recognition (Something-Something V1 and V2) and video captioning (MSVD \cite{chen2011collecting}). For fair comparison, during the training of MUFI, we remove the training videos that are duplicate of validation videos of these datasets. Two videos are judged to be duplicate if they share the same URL, or are visually very similar. More details for video deduplication will be given in the supplementary material.

\subsection{Implementation Details}
\textbf{Language embedding.} The 24-layer BERT \cite{devlin2018bert} is utilized as the language embedding model pre-trained on unannotated text from BooksCorpus \cite{zhu2015aligning} and Wikipedia. We extract the output of the last but one layer in BERT as a sequence of 1,024-dimensional word representations. The language representation is then produced by concatenating the features max-pooled and average-pooled over all words, resulting in 2,048-dimensional feature vector. Finally, the language feature is reduced to 256 dimension by PCA to construct the embedding space.

\textbf{Training strategy.} Our MUFI is implemented in Caffe \cite{jia2014caffe} and the weights are trained by SGD. We employ LGD-3D \cite{qiu2019learning} originated from the ImageNet pre-trained ResNet-50 \cite{he2015deep} as our backbone. During training, the dimension of input video clip is set as $16\times 112\times 112$, which is randomly cropped from the non-overlapped 16-frame clip resized with the short edge in $\left[ 128, 170 \right]$. Random horizontal flipping and color jittering are utilized for augmentation.

\textbf{Evaluation protocol.} We exploit two evaluation protocols, i.e., linear model and network fine-tuning. In the former protocol, we directly exploit the backbone learnt by MUFI as the feature extractor, and verify the frozen representation via linear classification. Specifically, we uniformly extract 20 clip-level representation and average all the clip-level features to obtain the video-level representation. A linear SVM is trained on the training set and evaluated on each validation set. In the latter one, the pre-trained model by MUFI serves as the network initialization for further fine-tuning in downstream tasks. At inference stage, we employ the three-crop strategy in \cite{feichtenhofer2019slowfast} that crops three $128\times 128$ regions from each video clip.

\begin{table}[!tb]
\centering
\scriptsize
\caption{\small Per-facet performance comparisons across video representations learnt on single/multiple facet(s) in different ways.}
\begin{tabularx}{0.47\textwidth}{X|c@{~~~~}c@{~~~~}c@{~~~~}c@{~~~~}c@{~~~~}c|c}
\toprule
\textbf{Method} & \textbf{act.} & \textbf{eve.} & \textbf{int.} & \textbf{spo.} & \textbf{obj.} & \textbf{sce.} & \textbf{avg} \\
\midrule
\multicolumn{8}{l}{Video Representation Learning via Classification} \\ \midrule
action-only (K700)    & 57.6 & 17.9 & 17.8 & 53.1 & 74.2 & 93.1 & 52.2 \\
event-only (MIT) & 40.4 & 26.0 & 16.3 & 46.1 & 68.6 & 95.3 & 48.7 \\
interaction-only (SSV2)            & 18.9 & 5.6  & 38.6 & 25.6 & 46.6 & 67.1 & 33.7 \\
sport-only (S1M)        & 34.8 & 11.6 & 10.4 & 67.3 & 66.6 & 88.5 & 46.7 \\
4facets     & 51.3 & 22.7 & 37.4 & 62.9 & 74.2 & 95.1 & 57.2 \\
4facets+multi-fc & 51.4 & 21.0 & 40.8 & 61.1 & 73.5 & 95.2 & 57.1 \\ \midrule
\multicolumn{8}{l}{Video Representation Learning via Visual-Semantic Embedding} \\ \midrule
intra$_{L2}$ & 52.3 & 23.2 & 32.0 & 63.8 & 73.2 & 95.1 & 56.6 \\
intra$_{nce}$ &  55.6 & 25.9 & 42.4 & 65.3 & 74.7 & 95.6 & 59.9 \\
intra$_{nce}$+inter$_{video}$ & 56.2 & 26.2 & 44.1 & 65.7 & 74.6 & 94.8 & 60.2 \\
intra$_{nce}$+inter$_{all}$ & 55.7 & 26.2 & 44.3 & 65.6 & 76.7 & 96.9 & 60.9 \\
\textbf{MUFI} (+attention) & \textbf{57.8} & \textbf{27.0} & \textbf{47.8} & \textbf{67.7} & \textbf{77.8} & \textbf{97.5} & \textbf{62.6} \\
\bottomrule
\end{tabularx}
\label{tab:methods}
\vspace{-0.2in}
\end{table}

\subsection{Evaluations on MUFI w.r.t Every Facet}
We first examine the effectiveness of MUFI for video representation learning with respect to every facet under linear model protocol from three perspectives: (1) video representations learnt on single/multiple facet(s) in different ways, (2) difference between MUFI and training/fine-tuning the network by considering one facet after another, and (3) comparisons with video representations extracted by off-the-shelf vision models.

\textbf{Video representations learnt on single/multiple facet(s) in different ways.} We compare the following runs for performance evaluation. The run of action-only/event-only/interaction-only/sport-only performs video representation learning via classification on Kinetics-700 (K700), Moments-In-Time (MIT), Something-Something V2 (SSV2) and Sports1M (S1M), respectively, which leverages only one facet for model training. The run, 4facets, exploits one fc layer to execute classification on the union of all annotated categories on four facets. A variant of 4facets, namely 4facets+multi-fc, separates the fc layer into 4 groups, each of which corresponds to the categories from one specific facet. Another line to learn video representation delves into visual-semantic embedding. The methods, intra$_{L2}$ and intra$_{nce}$, capitalize on intra-facet supervision of the four datasets. The former formulates the embedding learning as minimizing L2 distance of video-label pairs, while the latter exploits NCE loss in contrastive learning framework. The run of intra$_{nce}$+inter$_{video}$ and intra$_{nce}$+inter$_{all}$ further take the inter-facet supervision across four video datasets and four video datasets plus two image datasets into account for cross-view embedding learning.
MUFI is the proposal in the paper that involves attention mechanism to reflect the focus of each facet.

Table \ref{tab:methods} summarizes the result comparisons on the validation sets of different facets. Specifically, action-only/event-only/interaction-only/sport-only method achieves the highest accuracy on its corresponding facet of evaluations, but performs relatively poor on other facets. The results are expected as solely learning on one dataset biases video representation to only one facet of the training dataset. Directly treating the four datasets as a whole and implementing classification of 4facets and 4facets+multi-fc leads the accuracy by 4.9\%$\sim$23.5\% on average over single facet. The results basically verify the merit of multi-faceted learning to enhance video representation. Though intra$_{L2}$ and intra$_{nce}$ involve the utilization of intra-facet supervision, they are different in the way that intra$_{L2}$ exploits each video-label connection independently, and intra$_{nce}$ explores the influence across different connections. As indicated by the results, intra$_{nce}$ leads to the boost against intra$_{L2}$ by 3.3\%.

The run of intra$_{nce}$+inter$_{video}$ and intra$_{nce}$+inter$_{all}$ further employ the inter-facet supervision and improve the average accuracy from 59.9\% to 60.2\% and 60.9\%. Note that intra$_{nce}$+inter$_{video}$ explores the inter-supervision from the first four facets and emphasizes more on these facets. As a result, it is not surprising that intra$_{nce}$+inter$_{video}$ exhibits consistently better performances than intra$_{nce}$ on these facets but with slightly worse performances on the other two facets. In comparison, intra$_{nce}$+inter$_{all}$ is benefited from the inter-supervision among all the facets in both video and image datasets and outperforms intra$_{nce}$ across all facets. With the multi-attention projection, MUFI nicely balances the contribution of each facet for video representation learning and obtains a clear performance gain from 60.9\% to 62.6\%. More importantly, MUFI achieves the highest accuracy across all the facets, showing the potential of learning more discriminative and generalized features.

\begin{table}[!tb]
\centering
\scriptsize
\caption{\small Per-facet comparisons between MUFI and training/fine-tuning the network by considering one facet after another.}
\begin{tabularx}{0.48\textwidth}{X|c@{~~~~}c@{~~~~}c@{~~~~}c@{~~~~}c@{~~~~}c|c}
\toprule
\textbf{Method} & \textbf{act.} & \textbf{eve.} & \textbf{int.} & \textbf{spo.} & \textbf{obj.} & \textbf{sce.} & \textbf{avg} \\
\midrule
K700 & 57.6 & 17.9 & 17.8 & 53.1 & 74.2 & 93.1 & 52.2 \\
K700+MIT & 45.0 & \textbf{27.3} & 16.6 & 47.6 & 70.0 & 94.8 & 50.2 \\
K700+MIT+SSV2 & 21.7 & 9.1 & 39.7 & 28.0 & 43.7 & 64.2 & 34.4 \\
K700+MIT+SSV2+S1M & 35.7 & 18.2 & 12.3 & \textbf{68.8} & 69.2 & 88.2 & 48.7 \\ \midrule
\textbf{MUFI} & \textbf{57.8} & 27.0 & \textbf{47.8} & 67.7 & \textbf{77.8} & \textbf{97.5} & \textbf{62.6} \\
\bottomrule
\end{tabularx}
\label{tab:iterative}
\vspace{-0.2in}
\end{table}

\textbf{Difference between MUFI and training/fine-tuning the network by considering one facet after another.} One common way in practice for video representation learning on multiple datasets/facets is to pre-train the network on a large-scale video dataset and then fine-tune the network by other datasets one by one. We take K700 as the basis to pre-train the network and fine-tune the architecture successively with MIT, SSV2 and S1M. Table \ref{tab:iterative} details the changes of accuracy as the network fine-tuning proceeds with one dataset after another. An interesting observation is that the performance gain tends to be large on the facet of dataset the most recently used for fine-tuning while the performances on other facets learnt before that drop significantly. For instance, fine-tuning the network with S1M boosts up the accuracy on the facet of sport from 28\% to 68.8\%. In the meantime, the accuracy on the facet of action decreases from 57.6\% to 35.7\%. We speculate that this may be the result of local emphasis on the latest facet and the knowledge learnt before may be partially ``forgot.'' In contrast, MUFI explores the supervision among all the facets holistically and performs the representation learning on multiple datasets simultaneously. The results again demonstrate the advantage of our MUFI on learning generalized representation across all the facets.

\begin{table}[!tb]
\centering
\scriptsize
\caption{\small Per-facet comparisons with video representations extracted by off-the-shelf vision models.}
\begin{tabularx}{0.48\textwidth}{X@{~~~}|c@{~~~}c@{~~~}c@{~~~}c@{~~~}c@{~~~}c|c}
\toprule
\textbf{Method} & \textbf{act.} & \textbf{eve.} & \textbf{int.} & \textbf{spo.} & \textbf{obj.} & \textbf{sce.} & \textbf{avg} \\
\midrule
\multicolumn{8}{l}{Image-based Representation} \\ \midrule
ResNet-50 \cite{he2015deep} & 33.1 & 10.5 & 11.1 & 42.5 & \textbf{80.5} & 89.0 & 44.4 \\
SimCLR \cite{chen2020simple} (self-supervised) & 31.4 & 10.8 & 12.2 & 42.4 & 75.3 & 89.7 & 43.6\\
MoCo v2 \cite{he2020momentum} (self-supervised) & 32.0 & 11.2 & 12.1 & 42.7 & 79.4 & 90.0 & 44.5\\ \midrule
\multicolumn{8}{l}{Video-based Representation} \\ \midrule
C3D \cite{tran2015learning} & 24.6 & 8.6 & 8.2 & 60.1 & 57.6 & 83.9 & 40.5 \\
P3D ResNet-152 \cite{qiu2017learning} & 35.8 & 11.2 & 11.9 & 60.8 & 74.7 & 90.4 & 47.3 \\
R(2+1)D-34-IG65M \cite{ghadiyaram2019large} & 53.1 & 21.1 & 15.8 & 54.5 & 70.5 & 95.2 & 51.7 \\
R(2+1)D-34-IG65M+K400 \cite{ghadiyaram2019large} & 54.1 & 20.0 & 14.3 & 50.8 & 67.9 & 92.9 & 50.0 \\
SeCo \cite{yao2020seco} (self-supervised) & 34.3 & 11.8 & 11.5 & 45.2 & 63.7 & 88.7 & 42.5 \\ \midrule
\textbf{MUFI} & \textbf{57.8} & \textbf{27.0} & \textbf{47.8} & \textbf{67.7} & 77.8 & \textbf{97.5} & \textbf{62.6} \\
\bottomrule
\end{tabularx}
\label{tab:models}
\vspace{-0.2in}
\end{table}

\textbf{Comparisons with video representations extracted by off-the-shelf vision models.} Here, we include both image-based and video-based models learnt in supervised, weakly-supervised or self-supervised manner for comparison, and Table \ref{tab:models} lists the results. In general, ResNet-50, SimCLR and MoCo V2 all show good performances on the object facet, and MoCo V2 slightly outdoes ResNet-50 on average. Compared to C3D, P3D devises a deeper network and involves the pre-training on ImageNet, showing better performances across all the facets. R(2+1)D-34-IG65M model is trained on 65 million weakly-supervised social-media videos and hashtags, and performs fairly well with good generalization to all the facets. Further fine-tuning the model with K400, improves feature discrimination on the facet of action, but unfortunately, affects the generalization on other facets. SeCo, as a self-supervised model on video data, is superior to MoCo V2 on the motion-related facets, e.g., action, event and sport, but inferior to MoCo V2 on the appearance-related facets, e.g., object and scene. Note that because MUFI only leverages the inter-facet supervision from image data to strengthen the learning of object facet in video representation, the learning on object facet may not be utilized as fully as direct object classification in ResNet-50. As such, MUFI performs slightly worse than ResNet-50 in object facet. On other five facets, MUFI leads to significant improvements against all the off-the-shelf~models.

\subsection{Experimental Analysis}
\begin{figure}[!tb]
   \centering {\includegraphics[width=0.45\textwidth]{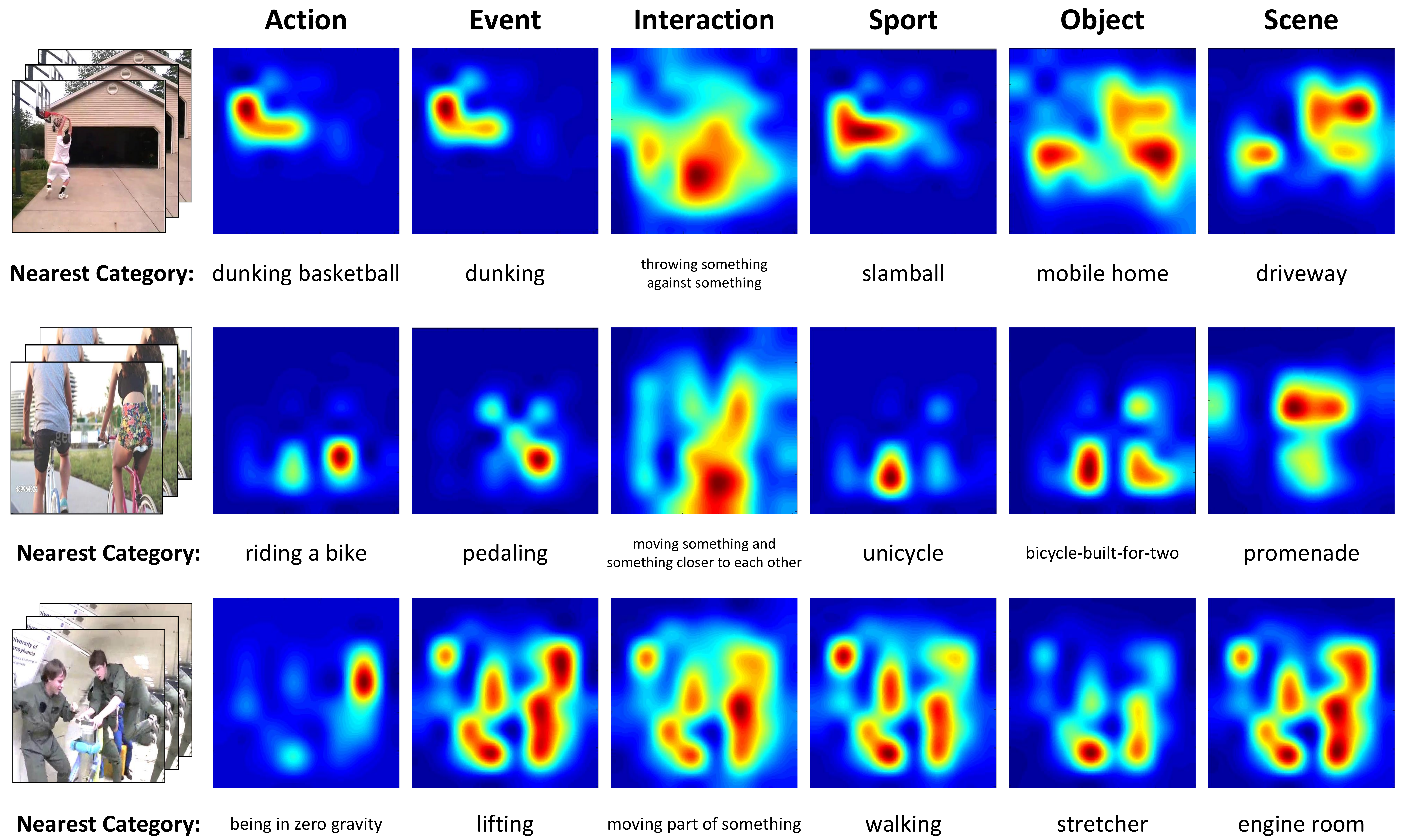}}
   \caption{\small Video examples with per-facet attention maps and the nearest label from each facet in embedding space.}
   \label{fig:attention}
   \vspace{-0.10in}
\end{figure}

\textbf{Visualization of attention map in each facet.} Multi-attention projection is uniquely devised in MUFI to locate the most indicative spatio-temporal regions from the viewpoint of each facet. Figure \ref{fig:attention} illustrates three video examples with the attention maps of all the six facets and the nearest label in terms of each facet in the embedding space. Overall, attention maps are expected to infer relevant video content in response to different facets. Taking the first video as an example, MUFI focuses on the regions around the Basket from action facet and pinpoints the regions of driveway with respect to the scene facet. Such multi-faceted integration equips video content understanding more comprehensive and endows video representation with more power.

\begin{figure}[!tb]
   \centering {\includegraphics[width=0.45\textwidth]{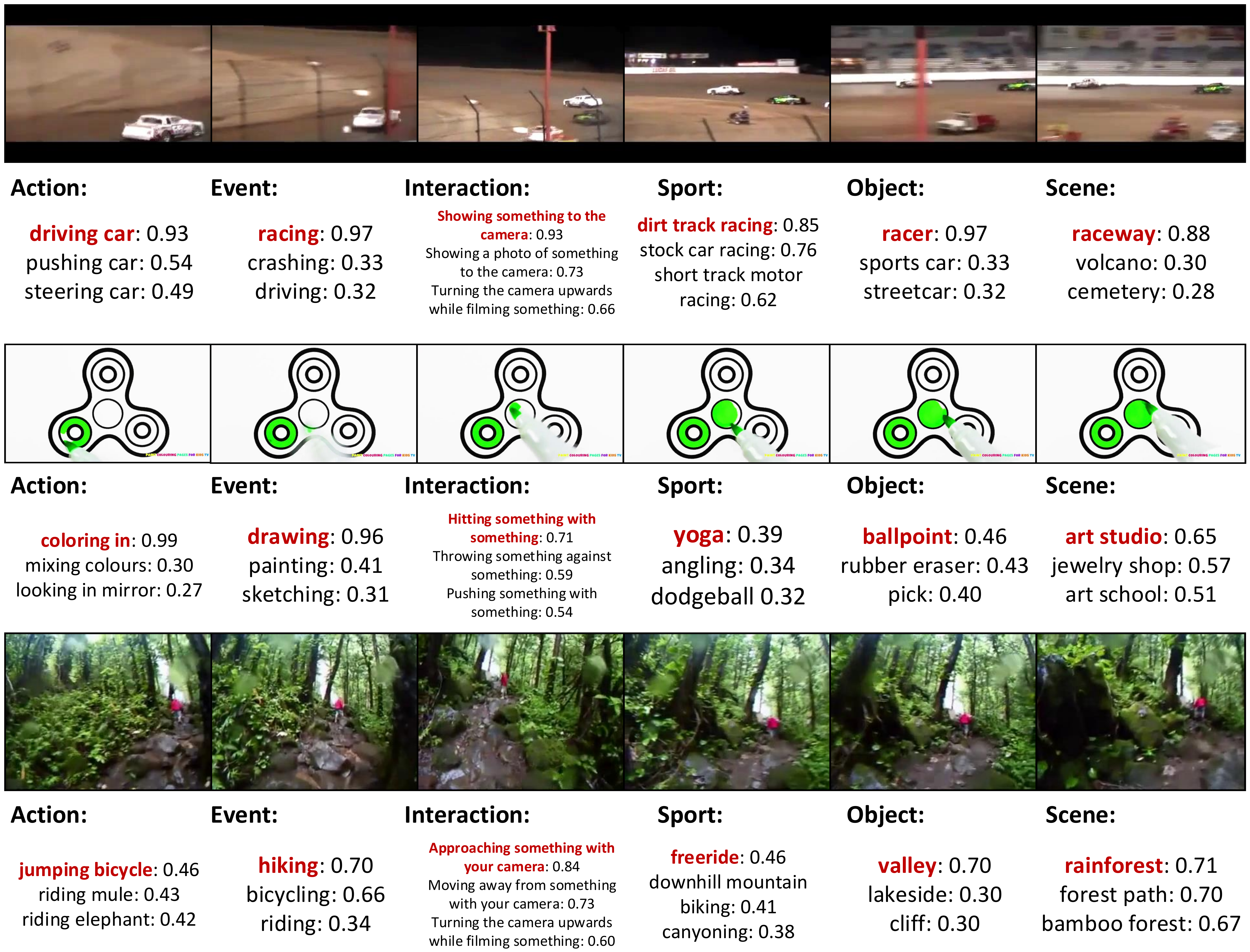}}
   \caption{\small Video examples with the top-3 relevant labels in the embedding space from every facet.}
   \label{fig:embedding}
   \vspace{-0.2in}
\end{figure}

\textbf{Examples of visual-semantic embedding.} Next, we qualitatively study the learnt visual-semantic embedding by MUFI. Figure \ref{fig:embedding} showcases three video clips with the top-3 relevant labels in the embedding space from every facet and each video clip is presented by six frames. As shown in the Figure, most labels are appropriate for describing the video content from different facets. For example, driving car in action facet, racing from event standpoint, racer from object perspective and raceway with respect to scene facet are all correlated with the first video and offer more dimensions of knowledge for video representation learning.

\begin{table}[!tb]
\centering
\scriptsize
\caption{\small Performance comparisons with the state-of-the-art methods with RGB input on UCF101 (3 splits) \& HMDB51 (3 splits).}
    \begin{tabularx}{0.48\textwidth}{l|c|c|CC}
        \toprule
       \textbf{Method} & \textbf{Backbone} & \textbf{Pre-train} & \textbf{UCF101} & \textbf{HMDB51}\\ \midrule
        \multicolumn{5}{l}{Linear/Fine-tuning Protocol} \\ \midrule
        I3D \cite{carreira2017quo} & BN-Inception & Kinetics-400  & 95.4 & 74.5\\
        ARTNet \cite{Wang2018AppearanceandRelationNF} & BN-Inception & Kinetics-400  & 94.3 & 70.9\\
        R(2+1)D \cite{tran2018closer} & ResNet-34  & Kinetics-400 &96.8 & 74.5\\
        S3D-G \cite{xie2018rethinking} & BN-Inception & Kinetics-400 &96.8 & 75.9\\
        STM \cite{Jiang2019STMSA} & ResNet-50 & Kinetics-400 &96.2  & 72.2\\
        LGD-3D \cite{qiu2019learning} & ResNet-101 & Kinetics-600 & 97.0 & 75.7 \\
        \textbf{MUFI+linear} & ResNet-50 & Multi-Faceted & 95.7 & 72.2 \\
        \textbf{MUFI+fine-tuning} & ResNet-50 & Multi-Faceted & \textbf{98.1} & \textbf{80.9} \\
       \midrule
        \multicolumn{5}{l}{Zero-shot Protocol} \\ \midrule
        SAOE \cite{mettes2017spatial} & GoogLeNet & ImageNet & 32.8 & -- \\
        CPD \cite{li2020learning} & 3D ResNet50 & Kinetics-210k & 39.9 & -- \\
        \textbf{MUFI} & ResNet-50 & Multi-Faceted & \textbf{56.3} & \textbf{31.0} \\
         \bottomrule
    \end{tabularx}
       \label{tab:hmdb_ucf}
\vspace{-0.1in}
\end{table}

\begin{table}[!tb]
\centering
\scriptsize
\caption{\small Performance comparisons with the state-of-the-art methods with RGB input on Kinetics-400.}
\begin{tabularx}{0.48\textwidth}{l@{~}|@{~}c@{~}|@{~}c@{~}|CC}
\toprule
\textbf{Method} & \textbf{Backbone} & \textbf{Pre-train} & \textbf{top-1} & \textbf{top-5} \\ \midrule
I3D \cite{carreira2017quo} & BN-Inception & ImageNet & 72.1 & 90.3 \\
R(2+1)D \cite{tran2018closer} & ResNet-34 & Sports1M & 74.3 & 91.4 \\
S3D-G \cite{xie2018rethinking} & BN-Inception & ImageNet & 74.7 & 93.4 \\
NL I3D \cite{wang2018non} & ResNet-101 & ImageNet & 77.7 & 93.3  \\
LGD-3D \cite{qiu2019learning} & ResNet-101 & ImageNet & 79.4 & 94.4 \\ \midrule
\multirow{3}{*}{SlowFast \cite{feichtenhofer2019slowfast}} & ResNet-50 & -- & 77.0 & 92.6 \\
& ResNet-101 & -- & 78.9 & 93.5 \\
& ResNet-101+NL & -- & 79.8 & 93.9 \\ \midrule
\multirow{2}{*}{R(2+1)D \cite{ghadiyaram2019large}} & ResNet-34 & IG65M & 79.6 & 93.9 \\
& ResNet-152 & IG65M & 81.3 & 95.1 \\ \midrule
\multirow{3}{*}{irCSN \cite{tran2019video}} & irCSN-152 & none & 76.8 & 92.5 \\
& irCSN-152 & Sports1M & 79.0 & 93.5 \\
& irCSN-152 & IG65M & 82.6 & 95.3 \\ \midrule
\textbf{MUFI+linear} & ResNet-50 & Multi-Faceted & 79.0 & 92.3 \\
\textbf{MUFI+fine-tuning} & ResNet-50 & Multi-Faceted & 81.1 & 95.1 \\
\textbf{MUFI+linear} & ResNet-101 & Multi-Faceted & 79.8 & 93.2 \\
\textbf{MUFI+fine-tuning} & ResNet-101 & Multi-Faceted & 82.3 & 95.3 \\
\bottomrule
\end{tabularx}
\label{tab:kinetics}
\vspace{-0.2in}
\end{table}

\subsection{Evaluations on Downstream Tasks}
We compare with several state-of-the-art methods on three downstream tasks: action recognition on UCF101, HMDB51 and Kinetics-400, interaction recognition on Something-Something V1/V2 (SS-V1/V2), and video captioning on MSVD. Table \ref{tab:hmdb_ucf} shows the comparisons on both UCF101 and HMDB51, and the performances of all the methods are reported with RGB input. MUFI+fine-tuning consistently outperforms other methods in the two datasets. In particular, MUFI+fine-tuning leads the accuracy by 1.1\% and 5.2\% over LGD-3D which uses a deeper backbone of ResNet-101. In addition to linear model and network fine-tuning protocols, we also include zero-shot protocol here, which performs classification by retrieving the nearest class based on the cosine distance between video clip and class labels in the embedding space. MUFI is again superior to SAOE and CPD. The results basically verify the generalization of the embedding space learnt by MUFI. Table \ref{tab:kinetics} summarizes the performances on Kinetics-400 dataset. Similarly, the performance gain of MUFI+fine-tuning against LGD-3D is 2.9\%, based on the same backbone of ResNet-101. MUFI+fine-tuning exhibits better performances than R(2+1)D-ResNet-152 with deeper network training on a much larger dataset of IG65M and performs comparable to irCSN trained on IG65M.

Table \ref{tab:ss} and Table \ref{tab:msvd} list the comparisons on SS-V1/V2 and MSVD for interaction recognition and video captioning, respectively. The top-1 accuracy of MUFI+fine-tuning achieves 51.2\% and 64.8\% on SS-V1 and SS-V2, making the absolute improvement over the best competitor STM by 0.5\% and 0.6\%. For video captioning, we exploit MUFI model to extract video representation and feed the representation into a transformer-based encoder-decoder structure to produce the sentence. As shown in the table, video representation learnt by MUFI leads to a CIDEr-D score boost of 6.3\% over the state-of-the-art model of ORG-TRL which fuses multiple video features extracted by Inception-ResNet-v2, C3D and Faster RCNN. This confirms the advantage of integrating multi-faceted information into video representation learning in our MUFI.

\begin{table}[!tb]
\centering
\scriptsize
\caption{\small Performance comparisons with state-of-the-art methods with RGB input on Something-Something V1/V2 (SS-V1/V2).}
\begin{tabularx}{0.48\textwidth}{l@{~}|@{~}c@{~}|CC|CC}
\toprule
\multirow{2}{*}{\textbf{Method}} & \multirow{2}{*}{\textbf{Pre-train}} & \multicolumn{2}{c|}{\textbf{SS-V1}} & \multicolumn{2}{c}{\textbf{SS-V2}} \\
& & Top-1 & Top-5 & Top-1 & Top-5 \\
\midrule
NL I3D+GCN \cite{Wang2018VideosAS} & ImageNet+K400 & 46.1 & 76.8 & -- & -- \\
TSM \cite{Lin2019TSMTS} & ImageNet+K400 & 47.2 & 77.1 & 63.4 & 88.5 \\
bLVNet-TAM \cite{Fan2019MoreIL} & ImageNet & 48.4 & 78.8 & 61.7 & 88.1 \\
ABM-C-in \cite{Zhu2019ApproximatedBM} & ImageNet & 49.8 & -- & 61.2 & -- \\
I3D+RSTG \cite{Nicolicioiu2019RecurrentSG} & ImageNet+K400 & 49.2 & 78.8 & -- & -- \\
GST \cite{Luo2019GroupedSA} & ImageNet & 48.6 & 77.9 & 62.6 & 87.9 \\
STDFB \cite{Martnez2019ActionRW} & ImageNet & 50.1 & 79.5 & -- & -- \\
STM \cite{Jiang2019STMSA} & ImageNet &  50.7 & 80.4 & 64.2 & 89.8 \\
\midrule
\textbf{MUFI+linear} & Multi-Faceted & 45.1 & 73.4 & 55.2 & 81.8 \\
\textbf{MUFI+fine-tuning} & Multi-Faceted & \textbf{51.2} & \textbf{81.0} & \textbf{64.8} & \textbf{90.0} \\
\bottomrule
\end{tabularx}
\label{tab:ss}
\vspace{-0.12in}
\end{table}

\begin{table}[!tb]
\centering
\scriptsize
\caption{\small Performance comparisons with state-of-the-art methods for video captioning on MSVD. (Evaluation metrics: BLEU@4 (\textbf{B@4}), METEOR (\textbf{M}), ROUGE-L (\textbf{R}) and CIDEr-D (\textbf{C})).}
\begin{tabularx}{0.48\textwidth}{@{~}l@{~}|@{~}c@{~}|CCCC}
\toprule
\textbf{Method} & \textbf{Backbone} & B@4 & M & R & C \\ \midrule
MARN \cite{pei2019memory} & ResNet-101+C3D & 48.6 & 35.1 & 71.9 & 92.2 \\
OA-BTG \cite{zhang2019object} & ResNet-200+Mask-RCNN & \textbf{56.9} & 36.2 & - & 90.6 \\
GRU-EVE \cite{aafaq2019spatio} & IncResV2+C3D+YOLO & 47.9 & 35.0 & 71.5 & 78.1 \\
MGSA \cite{chen2019motion} & IncResV2+C3D & 53.4 & 35.0 & - & 86.7 \\
POS+VCT \cite{hou2019joint} & IncResV2+C3D & 52.5 & 34.1 & 71.3 & 88.7 \\
ORG-TRL \cite{zhang2020object} & IncResV2+C3D+FasterRCNN & 54.3 & 36.4 & 73.9 & 95.2 \\
\midrule
\textbf{MUFI} & ResNet-50 & \textbf{56.9} & \textbf{37.4} & \textbf{74.2} & \textbf{101.5} \\
\bottomrule
\end{tabularx}
\label{tab:msvd}
\vspace{-0.15in}
\end{table}

\section{Conclusion}
We have presented MUlti-Faceted Integration (MUFI) framework, which explores various facets in videos to boost representation learning. Particularly, we study the problem from the viewpoint of integrating the knowledge from different datasets to represent multi-faceted information of a video. We first exploit the pre-trained language model to extract textual features of labels and build the semantic space across all the datasets as the embedding space. Each video representation is then mapped into the embedding space. Both intra-facet connection between a video and its own labels, and inter-facet predictions from other datasets are taken into account as multi-faceted supervision to optimize the visual-semantic embedding and eventually improve video representation. Experiments conducted on a union of four large-scale video datasets and two image datasets validate our proposal. More remarkably, our MUFI constantly obtains superior results over state-of-the-art methods in several downstream~tasks.

\textbf{Acknowledgments.} This work was supported by the National Key R\&D Program of China under Grant No. 2020AAA0108600. 

{\small
\bibliographystyle{ieee_fullname}
\bibliography{egbib}
}

\end{document}